\documentclass[conference]{IEEEtran}
\IEEEoverridecommandlockouts

\usepackage{paper}
\newcommand{\maxInRow}[1]{\bfseries #1}
\def\BibTeX{{\rm B\kern-.05em{\sc i\kern-.025em b}\kern-.08em
    T\kern-.1667em\lower.7ex\hbox{E}\kern-.125emX}}

\makeatletter
\def\ps@IEEEtitlepagestyle{%
  \def\@oddfoot{\mycopyrightnotice}%
}
\def\mycopyrightnotice{%
\fbox{\parbox{\dimexpr\textwidth-2\fboxsep-2\fboxrule\relax}{
\begin{minipage}{\textwidth-2\fboxsep-2\fboxrule}
  \footnotesize
  \textcopyright 2023 IEEE. Personal use of this material is permitted. Permission from IEEE must be obtained for all other uses, in any current or future media, including reprinting/republishing this material for advertising or promotional purposes, creating new collective works, for resale or redistribution to servers or lists, or reuse of any copyrighted component of this work in other works.
  \end{minipage}
}}
}
\makeatother

\begin{document}

\title{Towards Automated Regulatory Compliance Verification in Financial Auditing with \\ Large Language Models}

\author{\IEEEauthorblockN{
Armin Berger\IEEEauthorrefmark{1}\IEEEauthorrefmark{2}, Lars Hillebrand\IEEEauthorrefmark{1}\IEEEauthorrefmark{2}\thanks{* Both authors contributed equally to this research.}, David Leonhard\IEEEauthorrefmark{2}\IEEEauthorrefmark{3}, Tobias Deußer\IEEEauthorrefmark{2}\IEEEauthorrefmark{3}, Thiago Bell Felix de Oliveira\IEEEauthorrefmark{2}\IEEEauthorrefmark{3}\\Tim Dilmaghani\IEEEauthorrefmark{4}, Mohamed Khaled\IEEEauthorrefmark{4}, Bernd Kliem\IEEEauthorrefmark{4}
Rüdiger Loitz\IEEEauthorrefmark{4},\\
Christian Bauckhage\IEEEauthorrefmark{2}\IEEEauthorrefmark{3}, Rafet Sifa\IEEEauthorrefmark{2}\IEEEauthorrefmark{3}} \\
\IEEEauthorrefmark{2}\textit{Fraunhofer IAIS}, Sankt Augustin, Germany \\
\IEEEauthorrefmark{3}\textit{University of Bonn}, Bonn, Germany\\
\IEEEauthorrefmark{4}\textit{PricewaterhouseCoopers GmbH}, Düsseldorf, Germany
}
\IEEEpubid{\makebox[\columnwidth]{979-8-3503-2445-7/23/\$31.00 ©2023 IEEE \hfill} \hspace{\columnsep}\makebox[\columnwidth]{ }}


\maketitle

\begin{abstract}
The auditing of financial documents, historically a labor-intensive process, stands on the precipice of transformation. AI-driven solutions have made inroads into streamlining this process by recommending pertinent text passages from financial reports to align with the legal requirements of accounting standards. However, a glaring limitation remains: these systems commonly fall short in verifying if the recommended excerpts indeed comply with the specific legal mandates. Hence, in this paper, we probe the efficiency of publicly available Large Language Models (LLMs) in the realm of regulatory compliance across different model configurations. We place particular emphasis on comparing cutting-edge open-source LLMs, such as Llama-2, with their proprietary counterparts like OpenAI's GPT models. This comparative analysis leverages two custom datasets provided by our partner PricewaterhouseCoopers (PwC) Germany. We find that the open-source Llama-2 70 billion model demonstrates outstanding performance in detecting non-compliance or true negative occurrences, beating all their proprietary counterparts. Nevertheless,  proprietary models such as GPT-4 perform the best in a broad variety of scenarios, particularly in non-English contexts.
\end{abstract}


\section{Introduction}

Corporate financial disclosures in the form of financial statements provide critical insights into a firm's economic health and future trajectory.
These documents provide the public with detailed information on the financial stability, productivity, and profitability of a company, thus having a major influence on investment decisions made by external investors. Financial statements are documents that contain financial information of organizations such as assets, liabilities, and revenues. These documents are examined annually to check conformity with the relevant financial reporting framework, such as the International Financial Reports Standards (IFRS) and Germany's Handelsgesetzbuch (HGB). The examination process requires a lot of expert knowledge and manual analysis of lengthy financial texts. It includes tasks such as verifying the completeness, accuracy, valuation, consistency, classification, and readability of the reported information. The intricate nature of the IFRS and similar accounting standards exacerbate this challenge. Typically structured as an exhaustive list of checklist items, auditors are tasked with the responsibility of comparing relevant text passages from the financial document with each specific regulatory mandate. This necessitates the careful identification and correlation of text segments in the financial disclosure to the myriad stipulations in the accounting standard. With the advent of advanced Large Language Models (LLMs) like GPT-3.5-Turbo and GPT-4 \cite{openai2023gpt} showcasing impressive reasoning and text comprehension skills on various downstream tasks, we seek to explore their role in reshaping the auditing paradigm.

This paper builds upon our prior introduction of the Automated List Inspection (ALI) \cite{Sifa19} and the ZeroShotALI \cite{hillebrand2023zeroshotali} system to streamline the mapping between legal requirements and financial report segments. ZeroShotALI is a novel recommender system that leverages a state-of-the-art LLM in conjunction with a domain-specifically optimized transformer-based text-matching solution.

In this paper, we extend the capabilities of our current systems by investigating the potential of ``out-of-the-box" Language Models in evaluating the compliance of a legal requirement with a specified number of pertinent text passages extracted from financial documents. Our primary objectives encompass two key aspects: firstly, to evaluate the performance of open-source models in comparison to prominent proprietary models like GPT-4; and secondly, to analyze the impact of framing the problem through the utilization of prompts. 

Our motivation for exploring open-source models primarily stems from considerations related to cost-effectiveness and data privacy, both of which are pivotal concerns in the contemporary landscape of accounting and machine learning research.

In the following, we first review related work, before describing our modeling approach in Section~\ref{section:methodology}. In Section~\ref{section:experiments}, we outline our datasets, present our experiments, and discuss the results. Section~\ref{section:conclusion} then draws a conclusion and provides an outlook into conceivable future work.

\section{Related Work}
\label{sec:related}

The use of natural language processing (NLP) in the financial domain is an increasingly important field of research. To outline the contributions of other research that are related to this work, three areas are selected for distinctive comparison. First, the general field of financial NLP will be explored shortly. Then, research regarding the use of OpenAI's GPT line or the Llama-2 model, \cite{touvron2023llama}, for financial tasks will be captured. Last, general research on completeness or compliance checks, using NLP, will be presented. 

Within the extensive field of financial NLP, automated auditing is a sub-field that has been of interest to us for several years. The Automated List Inspection (ALI) tool \cite{Sifa19}, a supervised recommender system that ranks textual components of financial documents according to the requirements of established regulatory frameworks, such as IFRS, was introduced in 2019. Traditional NLP techniques like Tf-Idf, latent semantic indexing, neural networks, and logistic regression were employed to accomplish the ranking task. Combining the first and last methods resulted in the best performance. In \cite{Ramamurthy21}, we enhanced ALI by utilization of a pre-trained BERT (developed by \cite{Devlin19}) language model to encode text segments. A more general framework for this task was introduced by \cite{biesner2022zeroshot}. Further, to automatically verify the consistency of financial disclosures, a more detailed method for information extraction was needed. For this, we introduced KPI-Check, \cite{hillebrand2022towards}, also presented by \cite{hillebrand2022kpi}, a BERT-based system that utilizes a customized model for named entity and relation extraction. This tool automatically identifies and validates semantically equivalent key performance indicators in financial reports. The KPI extraction task, while previously focused on German documents, was also studied on an English dataset in \cite{deusser2022kpiedgar}, which was released together with the results. Combining text and table consistency checks in \cite{ali2023automatic}, we investigated financial reports using several pre-trained tabular models. The most important related work, which can be understood as a predecessor, is published in \cite{hillebrand2023zeroshotali}. There, we presented a novel recommender system for financial documents by employing a custom BERT-based model in conjunction with an LLM. In this work, we extend this approach by LLM-based completeness check given the recommended text segments.
Works by other researchers specializing in financial NLP include \cite{wu23bloomberggpt} who introduced \textsc{BloombergGPT}, employing Bloomberg's extensive data sources and evaluating the model's performance on financial tasks and general LLM benchmarks. FinBERT, introduced by \cite{araci2019finbert} and further employed and researched by \cite{yang2020finbert}, \cite{liu2021finbert}, \cite{arslan2021comparison}, and \cite{huang2022finbert} is another LLM intended for tasks specializing in financial language.
 
The GPT line of OpenAI is still relatively novel. Thus, there is limited research exploring its capabilities and restraints regarding the distinct use case of automated auditing. Additionally, as new versions are continuously being published, results obtained at different points in tine are comparable only to a limited degree. In the financial domain, there is work by \cite{cao2023bridging}, who qualitatively demonstrated GPT-4's effectiveness regarding the tasks of sentiment analysis, ESG analysis, corporate culture analysis, and Federal Reserve opinion analysis. 
\cite{neilson2023artificial} conducted a quantitative analysis using ChatGPT to generate financial recommendations for the Australian financial sector. The analysis revealed that ChatGPT was not effective in handling complex financial advice and needed additional professional guidance. Using GPT-3.5 and GPT-4 for validation, \cite{yue2023leveraging} proposed the Automated Financial Information Extraction framework to enhance the general ability of LLMs to extract KPIs from financial reports, finding significant average accuracy increases over a naive method. Last, \cite{callanan2023can} investigated the general capabilities of GPT-3.5 and GPT-4 in financial analysis, considering Zero-Shot, Chain-of-Thought, and Few-Shot scenarios, by prompting the models with mock exam questions from the Chartered Financial Analyst (CFA) program. 

\begin{figure}[t]
  \centering
  \includegraphics[width=0.95\linewidth]{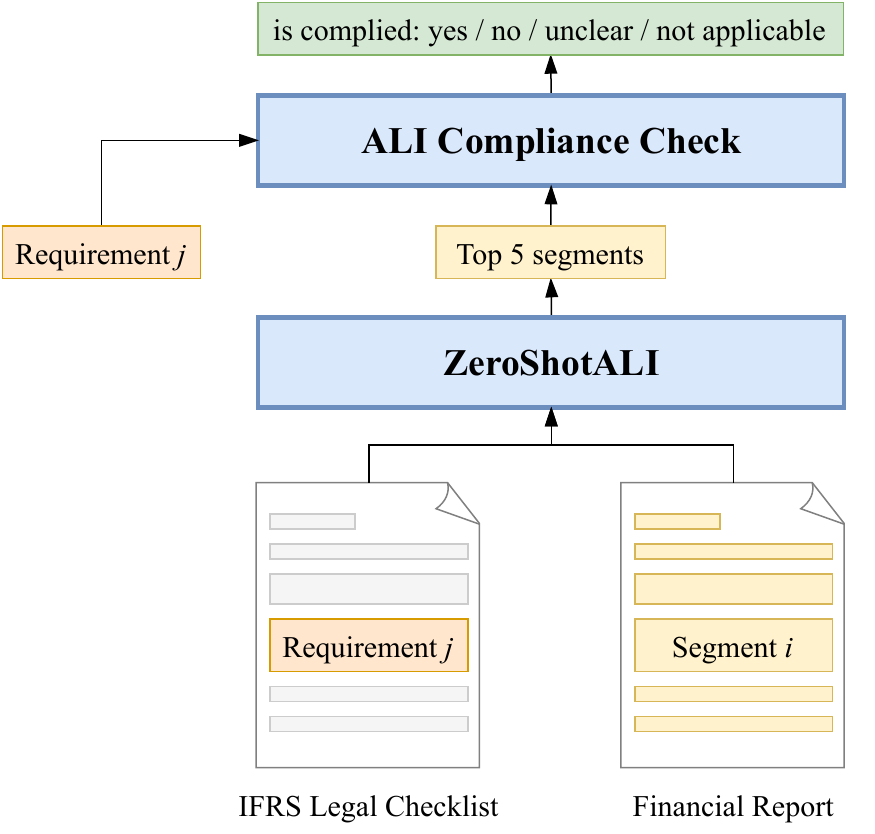}  
  \caption{Schematic visualization of the complete auditing pipeline combing our previous work, ZeroShotALI, an auditing-specific textual recommender system and the ALI compliance check system introduced in this work. While ZeroShotALI focuses on retrieving the top 5 relevant text passages per legal requirement, the compliance check system evaluates whether the retrieved passages comply with the provided requirement.}
  \label{fig:architecture}
\end{figure}


When it comes to the assurance of compliance and completion, which extends to the detection of contradictory statements, several works are noteworthy. The task of automated contradiction detection, using a transformer-based model, was studied by us in \cite{deusserCont2023}. Other work includes research on fraudulent statement detection by \cite{temponeras19}, who implemented a Deep Dense Multilayer Perceptron and compared it with other models such as Decision Trees, k-Nearest Neighbours, Support Vector Machines, and Logistic Regression. Additionally, \cite{zhu2021novel} put forth a novel capsule network to detect fraudulent activities in accounting reports. \cite{cao2018towards} employed a combined entity and relation extraction method to verify formulas in Chinese financial documents. Taking a look at the susceptibility of GPT-3 to manipulation of financial texts given the task of financial sentiment analysis, \cite{leippold2023sentiment} demonstrated the vulnerability of keyword-based approaches to adversarial attacks and with it arising the importance of robust models such as context-aware LLMs. Our research uniquely integrates a proven recommender system with an LLM to comprehensively verify the completeness of financial reports across varied reporting standards.

\section{Methodology} \label{section:methodology}

In this section, we briefly formulate the problem and motivate our modeling approach before turning to the in-depth analysis of our proposed architecture, which is visualized in Figure \ref{fig:architecture} and builds upon our prior research, ZeroShotALI.

In ZeroShotALI, we primarily focused on matching relevant text passages to legal mandates. However, the auditing workflow extends beyond this preliminary stage. Currently, auditors at PwC Germany employ tools that align each legal requirement within a financial reporting framework to its corresponding text passages. Subsequently, auditors must ascertain that these suggested passages not only pertain but also conform to the associated legal criterion. To streamline this process and enhance the efficacy of the auditing protocol, we propose the integration of LLMs to automatically validate the relevancy and compliance of these recommended text sections with their respective legal prerequisites. Further optimizing this process and alleviating the auditors' workload becomes increasingly critical in light of the expanding audit demands, driven by emerging regulatory frameworks such as the European Union's Corporate Sustainability Reporting Directive (CSRD).


\section{Experiments} \label{section:experiments}
In the following sections, we, first, introduce our two custom datasets, which are based on the International Financial Reports Standards (IFRS) and Germany's Handelsgesetzbuch (HGB). Secondly, we showcase how we evaluate the performance of different configurations. Following that we elaborate on what model and prompt configurations we deployed for each configuration. Finally, we present our results and discuss the implications of our findings on the auditing workflow.


\subsection{Data}

Our dataset\footnote{We are currently unable to publish the dataset and the accompanying Python code because both are developed and used in the context of an ongoing industrial project.} is based on 50 IFRS- and 50 HGB-compliant financial reports. In 2019 we introduced the Automated List Inspection (ALI) tool \cite{Sifa19}, a supervised recommender system that ranks text passages of financial documents according to the requirements of an auditing framework, such as IFRS or HGB. These machine-processed reports were then assessed by PwC auditors, who ensured that the relevant text segments in each report were correctly mapped to the corresponding IFRS or HGB accounting requirements. The task of annotating these machine-processed reports was evenly divided among three auditors, who were overseen by a senior auditor. Through several rounds of review, the generated labels for requirements were verified and fine-tuned. This was achieved by re-examining randomly chosen annotated samples and qualitatively assessing both false positives and negatives generated by the model.

Due to the size of the original datasets, we randomly sampled from both datasets. For the IFRS dataset, we sampled 10 financial reports, out of which we sub-sampled 100 requirements for our evaluation. For the HGB dataset, we randomly selected 3 reports, out of which we selected requirements for which at least two annotations existed. Since this paper assesses the ability of publicly available LLMs and does not train any models on domain-specific data, no splitting of the data was required. 

Due to the nature of the IFRS standard, the auditors annotated whether text passages complied with a requirement as either `yes' or `no'. For HGB all pairs of requirements and text passages were annotated as either `yes', `no', `unclear' or `not applicable'. Table  \ref{tab:comparison} shows the distribution of ground truth annotations for both datasets.

\begin{table}
    \centering
    \caption{Class distribution of ground truth values for IFRS and HGB.}
    \label{tab:comparison}
    \begin{tabular}{lcc}
    \toprule
    Label & IFRS & HGB \\
    \midrule
    Yes & 17 & 43 \\
    No & 82 & 28 \\
    Unclear & 1 & 23 \\
    Not Applicable & - & 26 \\
    \midrule
    Total & 100 & 120 \\
    \bottomrule
    \end{tabular}
    \begin{tablenotes}
        \vspace{2ex}
        \footnotesize
        \item \textit{Note: HGB Compliance Data was annotated by auditors at PwC Germany with `yes', `no', `unclear', or `not applicable', while IFRS Data was annotated with `yes', `no', and `unclear'.}
    \end{tablenotes}
\end{table}

\subsection{Evaluation Metrics}

We quantitatively evaluate our system's performance for each of the predicted classes by calculating precision, recall, and F$_1$ scores. The three metrics are defined as

\begin{align*}
    \label{eq:precision_and_sensitivity}
    \text{Precision}(c) &= \frac{|\text{TP}(c)|}{|\text{TP}(c)| + |\text{FP}(c)|} \\[10pt]
    \text{Recall}(c) &= \frac{|\text{TP}(c)|}{|\text{TP}(c)| + |\text{FN}(c)|}
\end{align*}

where $c \in \mathcal{C}$ represents the class (either ``yes", ``no", ``unclear", or ``not applicable"), $\text{TP}(c)$ represents the true positive observations, \( \text{FP}(c) \) represents the false positive observations, and \( \text{FN}(c) \) represents the false negative observations for class \( c \).

\begin{equation}
    \text{F}_{1}(c) = 2 \cdot \frac{\text{Precision}(c) \cdot \text{Recall}(c)}{\text{Precision}(c) + \text{Recall}(c)}.
\end{equation}

All scores are grouped by the prompt used to query a model and then averaged across the dataset to determine the overall performance per class. Additionally, we calculate the averages across all classes, `yes', `no', `unclear', or `not applicable', using a macro average, averaging the unweighted mean per class, and a micro average, averaging the support-weighted mean per class. Let \( \lvert \mathcal{C} \rvert \) denote the number of classes present per dataset. The two metrics are defined as
\begin{align*}
\text{Recall}_{\text{macro}} &= \frac{\sum_{i=1}^{\lvert \mathcal{C} \rvert} \text{Recall}(c_i)}{\lvert \mathcal{C} \rvert}  \\
\text{Recall}_{\text{micro}} &= \frac{\sum_{i=1}^{\lvert \mathcal{C} \rvert}  |\text{TP}(c_i)|}{\sum_{i=1}^{\lvert \mathcal{C} \rvert}  (|\text{TP}(c_i)| + |\text{FN}(c_i)|)}
\end{align*}

\subsection{Baselines}

In this study, we evaluate six state-of-the-art Large Language Models, comprising both open-source and proprietary architectures. Specifically, we include three variants of the open-source Llama-2 model\footnote{The concrete model IDs from Huggingface are: \href{https://huggingface.co/meta-llama/Llama-2-7b-chat-hf}{meta-llama/Llama-2-7b-chat-hf}, \href{https://huggingface.co/meta-llama/Llama-2-13b-chat-hf}{meta-llama/Llama-2-13b-chat-hf}, \href{https://huggingface.co/meta-llama/Llama-2-70b-chat-hf}{meta-llama/Llama-2-70b-chat-hf}.} \cite{touvron2023llama}, with sizes denoted as 7b, 13b, and 70b parameters, and three versions of the closed-source GPT architecture: GPT-3.5-Turbo, GPT-3.5-Turbo-16K, and GPT-4. Our rationale for juxtaposing open-source and proprietary models is twofold: economic considerations and data privacy issues. Additionally, the open-source nature of certain models offers the potential for fine-tuning, facilitating adaptability for specialized applications.

To ensure a controlled environment for model inference, we deployed a dedicated server equipped with an NVIDIA A100 80 GB GPU. We implemented an inference API analogous to the OpenAI API to facilitate on-demand access to the Llama-2 models. For experimental consistency, all models were subjected to identical prompts and parameters during the inference phase.
 
\subsection{Prompt Design}
\label{subsec:prompt_design}

Since all systems involve the querying of an LLM, we evaluate the impact of prompt design on model performance. The term prompt design encompasses how a task is presented to LLMs. Building on insights by \cite{liu2023pre} into the impact of prompt phrasing on LLM performance, our evaluation centers on two key factors: (1) task phrasing and (2) the structure of permitted model responses.

In our evaluation methodology, we have devised a specific task for all tested LLMs, involving the assessment of text passages from financial reports against regulatory accounting standards like IFRS or the German HGB. Through a process of trial and error and qualitative assessment, we have selected a total of eight prompts aimed at solving the above-stated task. Prompt performance was then quantified using metrics including Precision, Recall and F$_1$-Score per predicted class, as well as Micro and Macro F$_1$-Score averages across all classes (detailed in the Evaluation Metric section). Below we have added one exemplary prompt.




Exemplary  prompt:

\begin{tcolorbox}[breakable,notitle,boxrule=0pt,
boxsep=0pt,left=0.6em,right=0.6em,top=0.5em,bottom=0.5em,
colback=gray!10,
colframe=gray!10,
parbox=false]
System: You are an expert auditor with perfect knowledge of the IFRS accounting standard. 
You always answer truthfully whether a given regulatory requirement is fully complied with in the following line ids.

Answer with ``yes", if all sub-requirements are fully complied. Answer with ``no", 
if at least one of the sub-requirements is not fully complied.

Format your output complying to the following json schema:
\begin{verbatim}
{{"answer": <"yes"|"no">}}
\end{verbatim}
requirement: ``{requirement}"\\
document: ``{document}"
\end{tcolorbox}








The main two factors we discuss, are (1) task phrasing and (2) the structure of permitted model responses.

In terms of task phrasing, we used a variety of techniques such as chain of thought prompting, providing the model with an example prompt answer combination (also referred to as a one-shot prompt), asking the model for an explanation for their answer, and Tree-of-Thought prompting. Tree of Thought prompting as introduced by \cite{yao2023tree} and \cite{long2023large} (\url{https://www.promptingguide.ai/techniques/tot}) refers to the idea of enhancing LLM's ability to solve more elaborate problems through tree search via a multi-round conversation. Since this technique traditionally requires multiple LLM calls, the technique is costly and compute-intensive, thus not scaling well for commercial applications like ours. To overcome this issue, we have employed the prompting method from \cite{tree-of-thought-prompting} that adapts key elements from Tree-of-Thought frameworks to create a singular prompt that enables LLMs to assess intermediate ideas.

When examining the response structure, a noteworthy effect was observed. Outputs were categorized into two main formats: ``open-ended," allowing the model to provide explanations, and ``closed," constraining the model to return only `yes', `no', `unclear' or `not applicable'. Intriguingly, the ``closed'' format yielded superior performance compared to the ``open-ended" format. 

Due to the lengthy nature of each prompt, we summarize the differences between each evaluated prompt. The complete prompt definitions can be found in Appendix \ref{prompts}. 

\begin{tcolorbox}[breakable,notitle,boxrule=0pt,
boxsep=0pt,left=0.6em,right=0.6em,top=0.5em,bottom=0.5em,
colback=gray!10,
colframe=gray!10,
parbox=false]
{
\footnotesize

I. In-Out-Sub-Template:
\begin{itemize}
    \item Simple Yes/No/Unclear/Not applicable answer.
    \item Short, point-by-point answers without explanation.
    \item Formatting in JSON schema.
\end{itemize}
\vspace{10pt}

II. Cot-Sub-Template:
\begin{itemize}
    \item Chain of thought response leveraging intermediate explanations before the final answer.
    \item Specification of relevant line numbers from the document.
    \item No formatting in the JSON schema.
\end{itemize}
\vspace{10pt}

III. In-Out-Template:
\begin{itemize}
    \item Simplified version of In-Out-Sub.
    \item Direct Yes/No/Unclear/Not applicable response.
    \item Formatting in JSON Schema.
\end{itemize}
\vspace{10pt}

IV. Cot-Template:
\begin{itemize}
    \item Step-by-step response and explanation for each sub-request.
    \item Detailed explanation for each sub-request.
    \item No formatting in JSON schema.
\end{itemize}
\vspace{10pt}

V. In-Out-Tot-Template:
\begin{itemize}
    \item Same as In-Out, with the addition of tree-of-thought Prompting.
    \item Three experts give their opinion on the issue and come to a conclusion through majority voting.
\end{itemize}
\vspace{10pt}

VI. In-Out-Tot-One-Shot-Template:
\begin{itemize}
    \item Same as In-Out-Tot, with the addition of a one-shot example.
    \item In the one-shot example chosen, the text passage complies with the requirement.
\end{itemize}
\vspace{10pt}

VII. In-Out-One-Shot-Template:
\begin{itemize}
    \item Same as In-Out with the addition of a one-shot example.
    \item In the one-shot example chosen, the text passage complies with the requirement.
\end{itemize}
\vspace{10pt}

VIII. In-Out-One-Shot-No-Template:
\begin{itemize}
    \item Same as In-Out with the addition of a one-shot example.
    \item In the one-shot example chosen, the text passage does not comply with the requirement.
\end{itemize}
\vspace{10pt}
}
\end{tcolorbox}

\subsection{Evaluation and Results} \label{sec:results}

The goal of this paper is to assess whether current state-of-the-art  Large Language Models can be deployed in an auditing setting to validate the compliance of financial report passages with regulatory standards. To evaluate this, we selected six LLMs and deployed each on two custom financial auditing data sets. For each of these configurations, we ran the same eight selected prompts across all six LLMs, with the only exception being an answer adaptation for the open-source Llama-2 models. In the case of the LLM answering the task outside the scope of the predefined answer choices, the answer was cast as `invalid'. The three main questions we sought to answer by deploying a variety of different configurations are:

\begin{enumerate}
    \item[\textit{1)}] \textit{Performance Across Configurations:} Amongst all selected systems, which LLM and prompt configuration performed the best per class and across all classes?
    
    \item[\textit{2)}] \textit{Prompt Consistency Across Models:} Is prompt performance consistent across models? If yes, which prompts perform the best across all models and datasets?
    
    \item[\textit{3)}] \textit{Deploying LLM for Compliance:} How can we deploy an LLM-based compliance check system in a manner that saves auditors time while minimizing false negatives?
\end{enumerate}

\begin{table}[ht]
\centering
\caption{Micro F$_1$-Scores per model and dataset (HGB and IFRS) for the best performing prompt and the average over all prompts.}
\setlength{\tabcolsep}{4pt}
\begin{tabular}{lcccc}
\toprule
in \% & \multicolumn{2}{c}{HGB} & \multicolumn{2}{c}{IFRS} \\
\cmidrule(lr){2-3} \cmidrule(lr){4-5}
Model & Average & Best Prompt & Average & Best Prompt \\
\midrule
GPT-3.5-Turbo     & 41.83 & 46.15 & 66.68 & 77.56  \\
GPT-3.5-Turbo-16k & 40.48 & 46.15 & 49.92 & \maxInRow{77.58}  \\
GPT-4            & \maxInRow{59.31} & \maxInRow{75.60} & \maxInRow{71.65} & 77.05  \\
Llama2-7b        & 28.49 & 49.23 & 47.56 & 68.02  \\
Llama2-13b       & 24.81 & 47.34 & 48.01 & 65.58  \\
Llama2-70b       & 18.04 & 49.23 & 53.02 & 70.04  \\
\bottomrule
\end{tabular}
\label{micro_f1_per_model_dataset}
\end{table}

\subsubsection{Performance Across Configurations}
Evaluating the overall micro F$_1$-Score performance across both datasets, HGB and IFRS, the generally best-performing model is GPT-4. As can be seen in Table \ref{micro_f1_per_model_dataset}, the model achieves a micro F$_1$ score of 59.31\% on HGB and 71.65\% on IFRS data averaged across all prompts. A similar picture can be seen in the detailed analysis of comparing all prompt configurations for all models and datasets in Table \ref{tab:models_per_prompt} in Appendix \ref{prompt_evaluation}. Overall, we observed significantly worse performance on HGB data than on IFRS data across all models. This is likely attributed to the fact that most state-of-the-art LLMs are almost exclusively trained on an English text corpus, thus neglecting text understanding in other languages. The Llama-2 models for reference were trained on a 98\% English text corpus as stated in Meta's technical report \cite{touvron2023llama}.

A further notable finding is that the increasing parameter size in the open-source Llama-2 models did not translate into superior performance across all prompt types. Llama-2-70b for reference performed worse overall, considering micro F$_1$-Score performance across both the HGB and the IFRS dataset, than its significantly smaller counterpart Llama-2-7b.

\subsubsection{Prompt Consistency Across Models}
\begin{table}
 \caption{Best-performing prompt per model and dataset based on the Micro F$_1$-Score.}
  \centering
    \begin{tabular}{lll}
    \toprule
    Model & Prompt HGB & Prompt IFRS \\
    \midrule
    GPT-3.5-Turbo & I & VI \\
    GPT-3.5-Turbo-16k & I & I \\
    GPT-4 & VII & II \\
    Llama-2-7b & II & I \\
    Llama-2-13b & VI & III \\
    Llama-2-70b & VI & III \\
    \bottomrule
    \end{tabular}
  \label{tab:model_prompts}
\end{table}
Prompt performance does not appear to be consistent across models for our task. Despite the varying performance across models, we have noticed that prompt I. `In-Out-Sub-Template' achieved the best score in 4 out of 12 cases, when considering micro F$_1$-scores (see table \ref{tab:model_prompts}. These results indicate that a combination of keeping the prompt instructions brief and providing the model with examples to follow can improve the response quality. It is interesting to note that the more advanced prompting techniques such as chain of thought (II. and IV.)or trees of thought (VI. and VII.) did not induce a significant performance increase.

\begin{table}
 \caption{Results for Llama-2-70b in \%
  IFRS Data - Class `No'.}
  \centering
    \begin{tabular}{lccc}
    \toprule
    & \multicolumn{3}{c}{No} \\
    \cmidrule(lr){2-4}
    & Precision & Recall & F$_1$ \\
    \midrule
    I   & 75.00 & 41.25 & 53.23 \\
    II  & 80.22 & 91.25 & 85.38 \\
    III & 82.22 & 46.25 & 59.20 \\
    IV  & 82.86 & 72.50 & 77.33 \\
    V   & 81.58 & 77.50 & 79.49 \\
    VI  & 80.21 & \maxInRow{96.25} & \maxInRow{87.50} \\
    VII & 82.35 & 17.50 & 28.87 \\
    VIII& \maxInRow{85.00} & 21.25 & 34.00 \\
   \bottomrule
    \end{tabular}
  \label{tab:no_class_eval}
\end{table}

\subsubsection{Deploying LLMs for Compliance}
\begin{figure*}[t]
  \centering
  \includegraphics[width=0.8\linewidth]{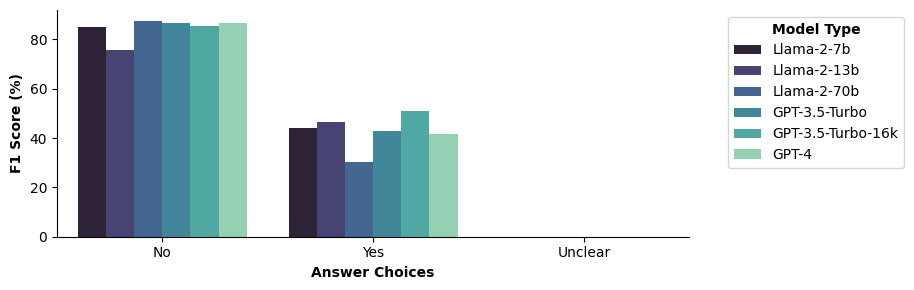}
  \caption{Grouped bar plot of F$_1$-Scores by model and answer choices on IFRS data.}
  \label{fig:results_ifrs}
\end{figure*}

\begin{figure*}[t]
  \centering
  \includegraphics[width=0.8\linewidth]{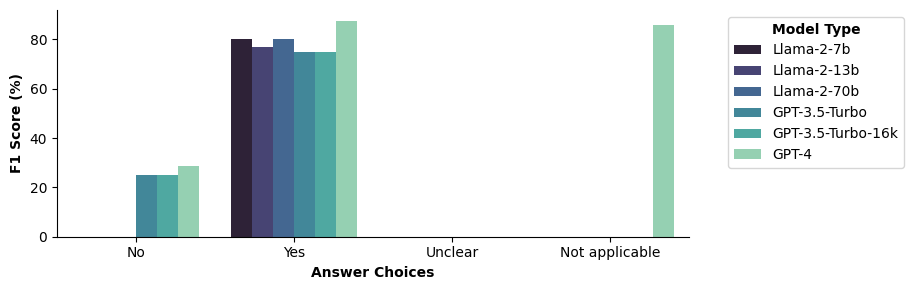}
  \caption{Grouped bar plot of F$_1$-Scores by model and answer choices on HGB data.} 
    {\footnotesize \textit{Note: Due to poor model performance in the German language, some LLMs were incapable of generating any machine-readable consistent outputs that are interpretable with a heuristic for some prompt formats, leading to some F$_1$-Scores being 0.}}
  \label{fig:results_hgb}
\end{figure*}
In collaboration with our partners at PwC Germany, we have determined that a compliance check system should avoid false positive predictions at all costs. Assuming that a text passage complying with a given auditing context is considered positive, a false negative prediction, in this case, would imply a financial report text passage falsely being classified as complying with a regulatory requirement. To avoid false positives, we want a model that has very high precision and recall for the class `No'. In our evaluations, we found that the best model for the `No' class in terms of F$_1$-Score is the open-source Llama-2-70b with a Precision of 80.21\%, Recall of 96.25\% and an F$_1$-Score of 87.50\% on IFRS data (see Table \ref{tab:no_class_eval}). This performance did not translate into a similar performance on HGB data, which is likely due to the German language.

Even though the `No' class is most important in practice, we also report the models' micro F$_1$ scores for the other classes in Figures \ref{fig:results_ifrs} for IFRS and \ref{fig:results_hgb} for HGB. It can be seen that for both datasets no model picked up on the `Unclear` class which might be explained by the models being overly confident in their `Yes' (is complied) and `No' (is not complied) answers. Also, the figures reveal that surprisingly for HGB the models perform best on the `Yes' class while performing significantly better on the `No' class for IFRS.


\section{Conclusion and Future Work}
\label{section:conclusion}

In our exploration to determine the suitability of publicly available Large Language Models (LLMs) in auditing settings for regulatory compliance verification, we have garnered several noteworthy insights.

Firstly, on the question of performance across configurations, the GPT-4 model demonstrated the most robust performance with an impressive micro F$_1$ score across the IFRS dataset given the complexity of the task. It is also crucial to underscore the challenge faced by LLMs when confronted with languages other than English. Our findings strongly suggest that the majority of LLMs are likely to exhibit sub-optimal performance on non-English datasets, like the German HGB, due to their primary training on English text corpora. This phenomenon was especially pronounced in the Llama-2 models.

Secondly, concerning prompt consistency across models, it became evident that there is no one-size-fits-all prompt. Different LLMs responded optimally to varied prompts, and more advanced prompting techniques did not consistently outperform their simpler counterparts. This underscores the necessity to custom-tailor prompts for different models and tasks to ensure peak performance. The success of brief prompts paired with exemplary one-shot instructions suggests the potential benefits of this approach.

Finally, in the context of deploying LLMs for compliance validation in financial reports, precision is of paramount importance. False positives can have significant repercussions in the auditing domain. Our collaboration with PwC Germany elucidated the importance of achieving high precision, especially for detecting non-compliance or `No' answers. Interestingly, the open-source Llama-2-70b model demonstrated outstanding performance in this regard on the IFRS dataset. To conclude, while LLMs hold immense promise in revolutionizing the auditing landscape, it is imperative to approach their deployment judiciously. Selecting the right model, tailoring the prompts, and acknowledging their shortcomings is imperative to ensuring they can be integrated as reliable components in an auditing pipeline. Given the results of our evaluation, we are currently not confident in the ability of ``out-of-the-box" Language Models to reliably assess the compliance of legal requirements.

In subsequent research, given the commendable performance of the open-source Llama-2-70b model in accurately predicting true negatives, it warrants additional investigation to explore the potential improvements achievable through further model training. Specifically, fine-tuning the model on comprehensive accounting compliance data may enhance its effectiveness and enable a reliable assessment of regulatory compliance. Opting for a dedicated open-source model also offers advantages in terms of data privacy, especially when hosted locally. Moreover, such an approach could be economically advantageous in the long run, considering the substantial costs associated with commercial LLM APIs, such as OpenAI's GPT-4.


\section{Acknowledgment}

This research has been funded by the Federal Ministry of Education and Research of Germany and the state of North-Rhine Westphalia as part of the Lamarr-Institute for Machine Learning and Artificial Intelligence, LAMARR22B.

\renewcommand*{\bibfont}{\footnotesize}
\printbibliography

\section{Appendix}
\label{appendix}

\subsection{Prompt evaluation}
\label{prompt_evaluation}

Table \ref{tab:models_per_prompt} shows the detailed evaluation of all prompt configurations per dataset and model based on the micro F$_1$ score.

\begin{table*}
\centering
\caption{Micro F$_1$-Scores for all models per dataset and prompt configuration.}
\begin{tabular}{llcccccc}
\toprule
Prompt & Dataset & GPT-3.5 & GPT-3.5-16K & GPT-4 & Llama-2-7b & Llama-2-13b & Llama-2-70b \\
\midrule
& HGB & 45.88 & 45.88 & \maxInRow{66.67} & 49.23 & 0.00 & 0.00 \\
\multirow{-2}{*}{I} & IFRS & 76.42 & \maxInRow{77.21} & 73.60 & 58.58 & 24.57 & 42.58 \\
\rowcolor{gray!10}
& HGB & 35.49 & 35.10 & 40.07 & \maxInRow{46.89} & 36.20 & 13.68 \\
\rowcolor{gray!10}
\multirow{-2}{*}{II} & IFRS & 69.69 & 10.32 & \maxInRow{77.05} & 33.13 & 52.83 & 70.04 \\
& HGB & 43.44 & 43.44 & \maxInRow{66.67} & 36.20 & 0.00 & 24.62 \\
\multirow{-2}{*}{III} & IFRS & 58.22 & 57.22 & \maxInRow{71.73} & 29.74 & 22.59 & 50.69 \\
\rowcolor{gray!10}
& HGB & 37.87 & 37.87 & 41.03 & 43.44 & 43.44 & \maxInRow{49.23} \\
\rowcolor{gray!10}
\multirow{-2}{*}{IV} & IFRS & 74.27 & 11.03 & \maxInRow{75.23} & 35.83 & 56.49 & 66.57 \\
& HGB & 43.96 & 33.57 & \maxInRow{53.85} & 38.46 & 0.00 & 13.68 \\
\multirow{-2}{*}{V} & IFRS & \maxInRow{72.00} & 60.77 & 71.37 & 35.68 & 65.41 & 69.65 \\
\rowcolor{gray!10}
& HGB & 37.87 & 37.87 & \maxInRow{75.60} & 0.00 & 46.89 & 43.08 \\
\rowcolor{gray!10}
\multirow{-2}{*}{VI} & IFRS & 77.56 & \maxInRow{77.58} & 66.38 & 68.02 & 65.58 & 70.00 \\
& HGB & 43.96 & 43.96 & \maxInRow{63.95} & 13.68 & 24.62 & 0.00 \\
\multirow{-2}{*}{VII} & IFRS & 70.04 & 70.04 & \maxInRow{70.13} & 65.82 & 61.62 & 24.76 \\
\rowcolor{gray!10}
& HGB & 46.15 & 46.15 & \maxInRow{66.67} & 0.00 & 47.34 & 0.00 \\
\rowcolor{gray!10}
\multirow{-2}{*}{VIII} & IFRS & 35.21 & 35.21 & \maxInRow{67.74} & 53.69 & 35.02 & 29.87 \\
\bottomrule
\end{tabular}
\\[2ex] 
\textit{Note: Due to the verbose nature of Llama-2 Models and their poor performance in the German language, Llama-2 was incapable of generating any machine-readable consistent outputs that are interpretable with a heuristic for some prompt formats. This leads to some Micro F$_1$-Scores being 0.}

\label{tab:models_per_prompt}
\end{table*}

\subsection{Prompt configurations}
\label{prompts}

The following features the list of evaluated prompts as outlined in Section \ref{subsec:prompt_design}. First, the prompts used for reports written under the IFRS are shown. Second, prompts used for German reports written for the HGB are listed.

\begin{enumerate}[label=\Roman*]
    \item In-Out-Sub-Template
    \begin{tcolorbox}[breakable,boxrule=0pt,boxsep=0pt,left=0.6em,right=0.6em,top=0.5em,bottom=0.5em,
    colback=gray!10,
    colframe=gray!10]
        \scriptsize
        ``System: You are an expert auditor with perfect knowledge of the IFRS accounting standard. 
        You always answer truthfully whether a given regulatory requirement is fully complied in the following line ids.
        
        Is the following IFRS sub-requirement fully complied in the following input document?
        Answer with `yes', if the sub-requirement is fully complied. Answer with `no', 
        if it is not fully complied.
        
        Format your output complying to the following json schema:
        \{\{`answer': \verb!<!`yes'\verb!|!`no'\verb!>!\}\}
        
        requirement: `\{requirement\}'
        document: `\{document\}'
        
        \{\{`answer':''
    \end{tcolorbox}
    \item Cot-Sub-Template
    \begin{tcolorbox}[breakable,boxrule=0pt,boxsep=0pt,left=0.6em,right=0.6em,top=0.5em,bottom=0.5em,
    colback=gray!10,
    colframe=gray!10]
        \scriptsize
        ``System: You are an expert auditor with perfect knowledge of the IFRS accounting standard. 
        You always answer truthfully whether a given regulatory requirement is fully complied in the following line ids.
        
        Is the following IFRS sub-requirement fully complied in the following input document?
        Think step by step: Explain whether it is fully complied and reference relevant line ids from
        the input document. Based on your explanation, determine whether the requirement is fully
        complied by answering with `yes' or `no'.
        
        requirement: `\{requirement\}'
        document: `\{document\}'
        
        \{\{`answer':''
    \end{tcolorbox}
    \item In-Out-Template
    \begin{tcolorbox}[breakable,boxrule=0pt,boxsep=0pt,left=0.6em,right=0.6em,top=0.5em,bottom=0.5em,
    colback=gray!10,
    colframe=gray!10]
        \scriptsize
        ``System: You are an expert auditor with perfect knowledge of the IFRS accounting standard. 
        You always answer truthfully whether a given regulatory requirement is fully complied in the following line ids.
        
        Answer with `yes', if all sub-requirements are fully complied. Answer with `no', 
        if at least one of the sub-requirements is not fully complied.
        
        Format your output complying to the following json schema:
        \{\{`answer': \verb!<!`yes'\verb!|!`no'\verb!>!\}\}
        
        requirement: `\{requirement\}'
        document: `\{document\}'
        
        \{\{`answer':''
    \end{tcolorbox}
    \item Cot-Template
    \begin{tcolorbox}[breakable,boxrule=0pt,boxsep=0pt,left=0.6em,right=0.6em,top=0.5em,bottom=0.5em,
    colback=gray!10,
    colframe=gray!10]
        \scriptsize
        ``System: You are an expert auditor with perfect knowledge of the IFRS accounting standard. 
        You always answer truthfully whether a given regulatory requirement is fully complied in the following line ids.
        
        Answer with `yes', if all sub-requirements are fully complied. Answer with `no', 
        if at least one of the sub-requirements is not fully complied.
        
        Format your output complying to the following json schema:
        \{\{`answer': \verb!<!`yes'\verb!|!`no'\verb!>!\}\}
        
        requirement: `\{requirement\}'
        document: `\{document\}'
        
        \{\{`answer':''
    \end{tcolorbox}
    \item In-Out-Tot-Template
    \begin{tcolorbox}[breakable,boxrule=0pt,boxsep=0pt,left=0.6em,right=0.6em,top=0.5em,bottom=0.5em,
    colback=gray!10,
    colframe=gray!10]
        \scriptsize
        ``System: Imagine three different experts in the field of auditing answering this question.
        Each expert has perfect knowledge of the IFRS accounting standard. All experts always answer
        truthfully whether a particular regulatory requirement in the following line numbers is
        is completely fulfilled.
        
        Each expert writes down 1 step of their thought process and shares it with the group.
        Then all experts move to the next step, and so on. Show each step and each expert's thinking process.
        If at any time an expert realizes he is wrong, he is eliminated.
        
        Answer with `yes', if all sub-requirements are fully complied. Answer with `no', 
        if at least one of the sub-requirements is not fully complied.
        
        Format your output complying to the following json schema:
        \{\{`answer': \verb!<!`yes'\verb!|!`no'\verb!>!\}\}
        
        requirement: `\{requirement\}'
        document: `\{document\}'
        
        \{\{`answer':''
    \end{tcolorbox}
    \item In-Out-Tot-One-Shot-Template
    \begin{tcolorbox}[breakable,boxrule=0pt,boxsep=0pt,left=0.6em,right=0.6em,top=0.5em,bottom=0.5em,
    colback=gray!10,
    colframe=gray!10]
        \scriptsize
        ``System: Imagine three different experts in the field of auditing answering this question.
        Each expert has perfect knowledge of the IFRS accounting standard. All experts always answer
        truthfully whether a particular regulatory requirement in the following line numbers is
        is completely fulfilled.
        
        Each expert writes down 1 step of their thought process and shares it with the group.
        Then all experts move to the next step, and so on. Show each step and each expert's thinking process.
        If at any time an expert realizes he is wrong, he is eliminated.
        
        Answer with `yes', if all sub-requirements are fully complied. Answer with `no', 
        if at least one of the sub-requirements is not fully complied.
        
        Format your output complying to the following json schema:
        \{\{`answer': \verb!<!`yes'\verb!|!`no'\verb!>!\}\}
        
        Requirement:`\{requirement\}'
        Document:`\{document\}`
        
        Example:
        
        Requirement: `Disclose the amount of receivables with a remaining maturity of more than one year; separately for each item reported.'
        Document: `544: All receivables have a remaining maturity of less than one year.
        
        Expert 1: `The requirement asks for the amount of receivables with a remaining maturity of more than one year. However, line 544 states that all receivables have a remaining maturity of less than one year. This seems to imply that there are no receivables with a remaining maturity of more than one year and the requirement has been met.'
        
        Expert 2: `I see Expert 1's point. Since the document indicates that all receivables have a remaining term of less than one year, we can assume that the requirement has been met.'
        
        Expert 3: `I agree with my colleagues. Although the document does not provide any other specific information, we can conclude from the context that the requirement has been met.'
        
        \{\{`answer': `yes'\}\}

        \{\{`answer':''
    \end{tcolorbox}
    \item In-Out-One-Shot-Template
    \begin{tcolorbox}[breakable,boxrule=0pt,boxsep=0pt,left=0.6em,right=0.6em,top=0.5em,bottom=0.5em,
    colback=gray!10,
    colframe=gray!10]
        \scriptsize
        ``System: You are an expert auditor with perfect knowledge of the IFRS accounting standard. 
        You always answer truthfully whether a given regulatory requirement is fully complied in the following line ids.
        
        Answer with `yes', if all sub-requirements are fully complied. Answer with `no', 
        if at least one of the sub-requirements is not fully complied.
        
        Format your output complying to the following json schema:
        \{\{`answer': \verb!<!`yes'\verb!|!`no'\verb!>!\}\}
        
        Requirement: `\{requirement\}'
        Document: `\{document\}'
        
        Example:
        
        Requirement: `Disclose the amount of receivables due in more than one year; separately for each item shown.'
        Document: `544: All receivables have a remaining maturity of less than one year.
        
        \{\{`answer': `yes'\}\}

        \{\{`answer':''
    \end{tcolorbox}
    \item In-Out-Tot-One-Shot-Template
    \begin{tcolorbox}[breakable,boxrule=0pt,boxsep=0pt,left=0.6em,right=0.6em,top=0.5em,bottom=0.5em,
    colback=gray!10,
    colframe=gray!10]
        \scriptsize
        ``System: You are an expert auditor with perfect knowledge of the IFRS accounting standard. 
        You always answer truthfully whether a given regulatory requirement is fully complied in the following line ids.
        
        Answer with `yes', if all sub-requirements are fully complied. Answer with `no', 
        if at least one of the sub-requirements is not fully complied.
        
        Format your output complying to the following json schema:
        \{\{`answer': \verb!<!`yes'\verb!|!`no'\verb!>!\}\}
        
        Requirement: `\{requirement\}'
        Document: `\{document\}'

        Example:
        
        Requirement: `Disclose the amount of receivables due in more than one year; separately for each item shown.'
        Document: `544: All receivables have a remaining maturity of less than one year.
        
        \{\{`answer': `no'\}\}

        \{\{`answer':''
    \end{tcolorbox}
\end{enumerate}
\begin{enumerate}[label=\Roman*]
    \item In-Out-Sub-Template (German)
    \begin{tcolorbox}[breakable,boxrule=0pt,boxsep=0pt,left=0.6em,right=0.6em,top=0.5em,bottom=0.5em,
    colback=gray!10,
    colframe=gray!10]
        \scriptsize
        ``System: Sie sind ein Experte im Bereich Wirtschaftsprüfung und haben perfekte Kenntnisse des 
        HGB-Bilanzierungstandards. Sie antworten immer wahrheitsgemäß, ob eine bestimmte behördliche Anforderung in den 
        folgenden Zeilennummern vollständig erfüllt ist.
        
        Ist die folgende HGB-Anforderung im unten genannten Dokument vollständig erfüllt?
        Antworten Sie mit:
        - `yes', wenn die Anforderung erfüllt ist,
        - `no', wenn die Anforderung nicht erfüllt ist,
        - `unclear', wenn die Erfüllung der Anforderung nicht beantwortet werden kann, da Kontextinformationen fehlen und
        - `not applicable', wenn die Anforderung für das Dokument nicht relevant ist.
        
        Formatieren Sie Ihre Ausgabe gemäß dem folgenden JSON-Schema:
        \{\{`answer': \verb!<!`yes'\verb!|!`no'\verb!|!`unclear'\verb!|!`not applicable'\verb!>!\}\}
        
        Anforderung: `\{requirement\}'
        Dokument: `\{document\}'
        
        \{\{`answer':''
    \end{tcolorbox}
    \item Cot-Sub-Template (German)
    \begin{tcolorbox}[breakable,boxrule=0pt,boxsep=0pt,left=0.6em,right=0.6em,top=0.5em,bottom=0.5em,
    colback=gray!10,
    colframe=gray!10]
        \scriptsize
        ``System: Sie sind ein Experte im Bereich Wirtschaftsprüfung und haben perfekte Kenntnisse des 
        HGB-Bilanzierungstandards. Sie antworten immer wahrheitsgemäß, ob eine bestimmte behördliche Anforderung in den 
        folgenden Zeilennummern vollständig erfüllt ist.
        
        Ist die folgende HGB-Teilanforderung im gegebenen Eingangsdokument vollständig erfüllt?
        Denke Schritt für Schritt: Erklären Sie, ob sie vollständig erfüllt ist, und geben Sie die relevanten Zeilennummern aus 
        dem vorhanden Dokument an. Basierend auf Ihrer Erklärung bestimmen Sie, ob die Anforderung vollständig erfüllt ist.
        Beenden Sie ihre Antwort mit
        - `yes', wenn die Anforderung erfüllt ist,
        - `no', wenn die Anforderung nicht erfüllt ist,
        - `unclear', wenn die Erfüllung der Anforderung nicht beantwortet werden kann, da Kontextinformationen fehlen und
        - `not applicable', wenn die Anforderung für das Dokument nicht relevant ist.
        
        Anforderung: `\{requirement\}'
        Dokument: `\{document\}'
        
        \{\{`answer':''
    \end{tcolorbox}
    \item In-Out-Template (German)
    \begin{tcolorbox}[breakable,boxrule=0pt,boxsep=0pt,left=0.6em,right=0.6em,top=0.5em,bottom=0.5em,
    colback=gray!10,
    colframe=gray!10]
        \scriptsize
        ``System: Sie sind ein Experte im Bereich Wirtschaftsprüfung und haben perfekte Kenntnisse des 
        HGB-Bilanzierungstandards. Sie antworten immer wahrheitsgemäß, ob eine bestimmte behördliche Anforderung in den 
        folgenden Zeilennummern vollständig erfüllt ist.
        
        Antworten Sie mit:
        - `yes', wenn die Anforderung erfüllt ist,
        - `no', wenn die Anforderung nicht erfüllt ist,
        - `unclear', wenn die Erfüllung der Anforderung nicht beantwortet werden kann, da Kontextinformationen fehlen und
        - `not applicable', wenn die Anforderung für das Dokument nicht relevant ist.
        
        Formatieren Sie Ihre Ausgabe gemäß dem folgenden JSON-Schema:
        \{\{`answer': \verb!<!`yes'\verb!|!`no'\verb!|!`unclear'\verb!|!`not applicable'\verb!>!\}\}
        
        Anforderung: `\{requirement\}'
        Dokument: `\{document\}'
        
        \{\{`answer':''
    \end{tcolorbox}
    \item Cot-Template (German)
    \begin{tcolorbox}[breakable,boxrule=0pt,boxsep=0pt,left=0.6em,right=0.6em,top=0.5em,bottom=0.5em,
    colback=gray!10,
    colframe=gray!10]
        \scriptsize
        ``System: Sie sind ein Experte im Bereich Wirtschaftsprüfung und haben perfekte Kenntnisse des 
        HGB-Bilanzierungstandards. Sie antworten immer wahrheitsgemäß, ob eine bestimmte behördliche Anforderung in den 
        folgenden Zeilennummern vollständig erfüllt ist.
        
        Denke Schritt für Schritt: Erkläre für jede Teilanforderung, ob sie vollständig erfüllt ist, und verweise in jeder Erklärung 
        auf die relevanten Zeilennummern aus dem gegebenen Dokument. Basierend auf deinen Erklärungen entscheide, 
        ob jede Teilanforderung vollständig erfüllt ist. 
        Beenden Sie ihre Antwort mit
        - `yes', wenn die Anforderung erfüllt ist,
        - `no', wenn die Anforderung nicht erfüllt ist,
        - `unclear', wenn die Erfüllung der Anforderung nicht beantwortet werden kann, da Kontextinformationen fehlen und
        - `not applicable', wenn die Anforderung für das Dokument nicht relevant ist.
        
        Anforderung: `\{requirement\}'
        Dokument: `\{document\}'
        
        \{\{`answer':''
    \end{tcolorbox}
    \item In-Out-Tot-Template (German)
    \begin{tcolorbox}[breakable,boxrule=0pt,boxsep=0pt,left=0.6em,right=0.6em,top=0.5em,bottom=0.5em,
    colback=gray!10,
    colframe=gray!10]
        \scriptsize
        ``System: Stellen Sie sich vor, drei verschiedene Experten im Bereich Wirtschaftsprüfung beantworten diese Frage.
        Jeder Experte hat perfekte Kenntnisse des HGB-Bilanzierungstandards. Alle Experten antworten immer
        wahrheitsgemäß, ob eine bestimmte behördliche Anforderung in den folgenden Zeilennummern
        vollständig erfüllt ist.
        
        Jeder Experte schreibt 1 Schritt seines Denkprozesses nieder und teilt ihn mit der Gruppe.
        Dann gehen alle Experten zum nächsten Schritt über, usw. .Zeige jeden Schritt und den Denkprocess jedes Expereten.
        Wenn ein Experte zu irgendeinem Zeitpunkt feststellt, dass er falsch liegt, scheidet er aus.

        Geben Sie bei einer Mehrheitsabstimmung unter den Experten nur eine Antwort in diesem Format zurück:
        - `yes', wenn die Anforderung erfüllt ist,
        - `no', wenn die Anforderung nicht erfüllt ist,
        - `unclear', wenn die Erfüllung der Anforderung nicht beantwortet werden kann, da Kontextinformationen fehlen und
        - `not applicable', wenn die Anforderung für das Dokument nicht relevant ist.
        
        Formatieren Sie Ihre Ausgabe gemäß dem folgenden JSON-Schema:
        \{\{`answer': \verb!<!`yes'\verb!|!`no'\verb!|!`unclear'\verb!|!`not applicable'\verb!>!\}\}
        
        Anforderung: `\{requirement\}'
        Dokument: `\{document\}'
        
        \{\{`answer':''
    \end{tcolorbox}
    \item In-Out-Tot-One-Shot-Template (German)
    \begin{tcolorbox}[breakable,boxrule=0pt,boxsep=0pt,left=0.6em,right=0.6em,top=0.5em,bottom=0.5em,
    colback=gray!10,
    colframe=gray!10]
        \scriptsize
        ``System: Stellen Sie sich vor, drei verschiedene Experten im Bereich Wirtschaftsprüfung beantworten diese Frage.
        Jeder Experte hat perfekte Kenntnisse des HGB-Bilanzierungstandards. Alle Experten antworten immer
        wahrheitsgemäß, ob eine bestimmte behördliche Anforderung in den folgenden Zeilennummern
        vollständig erfüllt ist.
        
        Jeder Experte schreibt 1 Schritt seines Denkprozesses nieder und teilt ihn mit der Gruppe.
        Dann gehen alle Experten zum nächsten Schritt über, usw. .Zeige jeden Schritt und den Denkprocess jedes Expereten.
        Wenn ein Experte zu irgendeinem Zeitpunkt feststellt, dass er falsch liegt, scheidet er aus.

        Geben Sie bei einer Mehrheitsabstimmung unter den Experten nur eine Antwort in diesem Format zurück:
        - `yes', wenn die Anforderung erfüllt ist,
        - `no', wenn die Anforderung nicht erfüllt ist,
        - `unclear', wenn die Erfüllung der Anforderung nicht beantwortet werden kann, da Kontextinformationen fehlen und
        - `not applicable', wenn die Anforderung für das Dokument nicht relevant ist.
        
        Formatieren Sie Ihre Ausgabe gemäß dem folgenden JSON-Schema:
        \{\{`answer': \verb!<!`yes'\verb!|!`no'\verb!|!`unclear'\verb!|!`not applicable'\verb!>!\}\}
        
        Anforderung: `\{requirement\}'
        Dokument: `\{document\}'
        
        Example:
        
        Anforderung: `Angabe des Betrags der Forderungen mit einer Restlaufzeit von mehr als einem Jahr; gesondert für jeden ausgewiesenen Posten'
        Dokument: `544: Sämtliche Forderungen haben eine Restlaufzeit von weniger als einem Jahr.
        
        Experte 1: `Die Anforderung verlangt die Angabe des Betrags der Forderungen mit einer Restlaufzeit von mehr als einem Jahr. In Zeile 544 wird jedoch angegeben, dass alle Forderungen eine Restlaufzeit von weniger als einem Jahr haben. Das scheint zu implizieren, dass es keine Forderungen mit einer Restlaufzeit von mehr als einem Jahr gibt und die Anforderung erfüllt worden ist.'
        
        Experte 2: `Ich sehe den Punkt von Experte 1. Da das Dokument angibt, dass alle Forderungen eine Restlaufzeit von weniger als einem Jahr haben, können wir davon ausgehen, dass die Anforderung erfüllt worden ist.'
        
        Experte 3: `Ich stimme meinen Kollegen zu. Obwohl das Dokument keine weiteren spezifischen Informationen enthält, können wir aus dem Kontext schließen, dass die Anforderung erfüllt worden ist.'
        
        \{\{`answer': `yes'\}\}
        
        \{\{`answer':''
    \end{tcolorbox}
    \item In-Out-One-Shot-Template (German)
    \begin{tcolorbox}[breakable,boxrule=0pt,boxsep=0pt,left=0.6em,right=0.6em,top=0.5em,bottom=0.5em,
    colback=gray!10,
    colframe=gray!10]
        \scriptsize
        ``System: Sie sind ein Experte im Bereich Wirtschaftsprüfung und haben perfekte Kenntnisse des 
        HGB-Bilanzierungstandards. Sie antworten immer wahrheitsgemäß, ob eine bestimmte behördliche Anforderung in den 
        folgenden Zeilennummern vollständig erfüllt ist.
        
        Antworten Sie mit:
        - `yes', wenn die Anforderung erfüllt ist,
        - `no', wenn die Anforderung nicht erfüllt ist,
        - `unclear', wenn die Erfüllung der Anforderung nicht beantwortet werden kann, da Kontextinformationen fehlen und
        - `not applicable', wenn die Anforderung für das Dokument nicht relevant ist.
        
        Formatieren Sie Ihre Ausgabe gemäß dem folgenden JSON-Schema:
        \{\{`answer': \verb!<!`yes'\verb!|!`no'\verb!|!`unclear'\verb!|!`not applicable'\verb!>!\}\}
        
        Anforderung: `\{requirement\}'
        Dokument: `\{document\}'
        
        Example:
        
        Anforderung: "Angabe des Betrags der Forderungen mit einer Restlaufzeit von mehr als einem Jahr; gesondert für jeden ausgewiesenen Posten"
        Dokument: "544: Sämtliche Forderungen haben eine Restlaufzeit von weniger als einem Jahr.
        
        \{\{`answer': `yes'\}\}
        
        \{\{`answer':''
    \end{tcolorbox}
    \item In-Out-Tot-One-Shot-Template (German)
    \begin{tcolorbox}[breakable,boxrule=0pt,boxsep=0pt,left=0.6em,right=0.6em,top=0.5em,bottom=0.5em,
    colback=gray!10,
    colframe=gray!10]
        \scriptsize
        ``System: Sie sind ein Experte im Bereich Wirtschaftsprüfung und haben perfekte Kenntnisse des 
        HGB-Bilanzierungstandards. Sie antworten immer wahrheitsgemäß, ob eine bestimmte behördliche Anforderung in den 
        folgenden Zeilennummern vollständig erfüllt ist.
        
        Antworten Sie mit:
        - `yes', wenn die Anforderung erfüllt ist,
        - `no', wenn die Anforderung nicht erfüllt ist,
        - `unclear', wenn die Erfüllung der Anforderung nicht beantwortet werden kann, da Kontextinformationen fehlen und
        - `not applicable', wenn die Anforderung für das Dokument nicht relevant ist.
        
        Formatieren Sie Ihre Ausgabe gemäß dem folgenden JSON-Schema:
        \{\{`answer': \verb!<!`yes'\verb!|!`no'\verb!|!`unclear'\verb!|!`not applicable'\verb!>!\}\}
        
        Anforderung: `\{requirement\}'
        Dokument: `\{document\}'
        
        Example:
        
        Anforderung: `Angabe des Betrags der Forderungen mit einer Restlaufzeit von mehr als einem Jahr; gesondert für jeden ausgewiesenen Posten'
        Dokument: `544: Sämtliche Forderungen haben eine Restlaufzeit von weniger als einem Jahr.
        
        \{\{`answer': `yes'\}\}
        
        \{\{`answer':''
    \end{tcolorbox}
\end{enumerate}

\end{document}